\documentclass[10pt]{cai}
\graphicspath{{figs/}}

\usepackage{adjustbox}

\begin{document}

\volumeheader{36}{0}
\begin{center}

    \title{Meta-GCN: A Dynamically Weighted Loss Minimization Method for Dealing with the Data Imbalance in Graph Neural Networks}
  \maketitle

  \thispagestyle{empty}

  \begin{tabular}{cc}
    Mahdi Mohammadizadeh\upstairs{\affilone}, Arash Mozhdehi\upstairs{\affilone}, Yani Ioannou\upstairs{\affilone}, Xin Wang\upstairs{\affilone,*}
   \\[0.25ex]
   {\small \upstairs{\affilone} University of Calgary} \\

  \end{tabular}
\end{center}

\begin{abstract}

Although many real-world applications, such as disease prediction, and fault detection suffer from class imbalance, most existing graph-based classification methods ignore the skewness of the distribution of classes; therefore, tend to be biased towards the majority class(es). 
Conventional methods typically tackle this problem through the assignment of weights to each one of the class samples based on a function of their loss, which can lead to over-fitting on outliers. 
In this paper, we propose a meta-learning algorithm, named Meta-GCN, for adaptively learning the example weights by simultaneously minimizing the unbiased meta-data set loss and optimizing the model weights through the use of a small unbiased meta-data set.
Through experiments, we have shown that Meta-GCN outperforms state-of-the-art frameworks and other baselines in terms of accuracy, the area under the receiver operating characteristic (AUC-ROC) curve, and macro F1-Score for classification tasks on two different datasets. 

\end{abstract}

\copyrightnotice

\section{Introduction}
\label{intro}
Graph-based data structures are ubiquitously used for modeling the pair-wise relation between the entities in a variety of real-world applications, including social networks \cite{tong2020stratlearner}, citation networks \cite{khan2019multi}, and protein-protein interactions \cite{zaki2021identifying}. 
Graph structures are effectively capable of describing the complex relationship between the objects, i.e. nodes, through edges. 
Besides, Graph-based representation is an effective method for feature dimensionality reduction \cite{zhang2018feature, 4016549}. 
GNNs, as powerful tools for representational learning on graph-structured data, have attracted increasing attention in recent years. 
GNNs are used for effective deep representational learning to perform graph analysis for tasks such as node classification, link prediction, and clustering in Euclidean and non-Euclidean domains \cite{zhou2020graph}.
Among the proposed methods for learning representations on graphs, 
GCNs proposed by Kipf et al. \cite{welling2016semi} proved to be a simple and effective GNN model. 
This model is able to learn hidden representations comprising both node features and local graph structure while scaling linearly relative to the number of edges in the given graph. 
Most classification algorithms, in GNNs, tend to minimize the average loss over all training examples which produces reasonable outcomes for class-balanced datasets. 
However, various real-world classification tasks, such as disease prediction \cite{kazi2019inceptiongcn}, fraud detection \cite{jiang2019anomaly}, and fault detection \cite{chen2019fault} manifest highly-skewed class distribution.
In settings that exhibit such class imbalance, these methods favor the majority classes while neglecting the minority ones. In addition, over-smoothing, which is a general issue in the GNN, can be aggravated in the case of data imbalance, as minority class nodes' representations become similar to that of majority ones \cite{10.1145/1122445.1122456}. 
Hence, these classifiers cannot well-differentiate the boundary between the minority and majority classes.
However, in applications like disease prediction, fault detection, and fraud detection, minorities are important, and classifying them correctly is crucial.
 Alternative to the existing methods, aiming to alleviate class imbalance for node classification on graph-structured data, we proposed an algorithm-based method, called Meta-GCN, that uses a meta-learning algorithm that adaptively assigns weights to the training examples in a way that minimizes the aggregated loss of an unbiased example set sampled from a meta-data set. 
 
 Our contributions are threefold: 
\begin{enumerate}
  \item Proposing a general-purpose online re-weighting algorithm for semi-supervised node classification in graphs that learns weighted loss function which is parameterized by weights that are learned in a meta-learning manner through the use of a small unbiased meta-data set. 
  \item Proposing a novel graph-based sampling method from the designated portion of the dataset to construct the meta-data set. 
  \item Through our experiment, we demonstrated that Meta-GCN outperforms
state-of-the-art frameworks and other baselines in terms of accuracy, the area under the
receiver operating characteristic (AUC-ROC) curve, and macro F1-Score for classification tasks
on two different datasets.
\end{enumerate}

\section{Related Work}
\label{related_work}
Proposed methods for addressing class imbalance issues on both graph and non-graph data fall into data-level, algorithm-level, or hybrid methods. 
Over-sampling and under-sampling are among the data-level solutions. While over-sampling methods aim to balance the ratio of classes by having more examples from minority classes in training, under-sampling solutions remedy the disproportion by removing the majority class(es) instances. 
Over-sampling with replicating the examples is known to tend to overfit; hence, a synthetic minority over-sampling technique (SMOTE) \cite{chawla2002smote} has been proposed to overcome this issue by generating synthetic minority examples through interpolating neighboring minority examples. 
To improve SMOTE, several variants of it have been proposed. However, over-sampling methods are error-prone by synthesizing examples close to the boundary.
Therefore, under-sampling methods are often preferred \cite{huang2019deep}, however, removing examples can lead to losing valuable instances required for discrimination, and as a result poor generalization. 
On the other hand, re-weighting methods, as algorithm-level solutions, aim to minimize a weighted loss on the training samples by assigning weights in a manner that pays more attention to minority examples. 
AdaBoost \cite{freund1996experiments} is a re-weighting-based approach that creates an ensemble of classifiers and assigns higher weights to misclassified instances. 
In \cite{6252738}, authors proposed improvement to AdaBoost by a hybrid method that first combines it with over-sampling and then uses an optimization algorithm that further tunes the class-specific weights. 
\cite{ren2018learning} proposed an algorithm that re-weights the training examples online based on gradient direction using a small clean validation set. 
Alternatively, in \cite{9920039}, authors proposed a generic method for evaluating the semantic completeness of datasets. 
Since these methods assume that the data is i.i.d they are not applicable to graph-based representations. 
As GNN-based classifiers are newly emerged, few works have instigated solving class imbalance issues in this area. 
GraphSMOTE \cite{zhao2021graphsmote}, a data-level approach for graph-structured data, has been proposed that generates the synthetic data in the embedding space. 
These methods proved to be error-prone. 
DR-GCN \cite{9766044} is an algorithm-level approach that uses class-conditioned adversarial regularizers to overcome the imbalance in graphs. 
This approach, as a hybrid method with adversarial training, is shown to suffer from instability and is unable to scale to large graphs.
Alternatively, Meta-GCN unlike most of the proposed approaches that deal with data imbalance is for semi-supervised node classification in graphs and we do not assume that the data is i.i.d. 
Unlike GraphSMOTE, our proposed method is an algorithm-level approach and does not generate synthetic data.

\section{Method}
\label{method}
In this section, we introduce the definitions, state the problem, and present our solution.
Assume a feature matrix $\mathcal{X} \in \mathbb{R} ^ {N \times F}$, with an imbalanced distribution of labels, where $N$ and $F$ are respectively the numbers of examples in the dataset and the number of features assigned to them.
Let $A \in \{0, 1\} ^ {N \times N}$ be a binary adjacency matrix, showing the connectivity between the examples in the dataset. 
For given examples $i$ and $j$ the entry $A_{i,j}$ is 1 iff the examples are connected and 0 otherwise. 
Given the feature matrix $\mathcal{X}$ and the adjacency matrix $A$, we construct the undirected and unweighted graph $\mathcal{G} = (V, E, \mathcal{X})$, where $V = \{1,\dots, N\}$ stands for the set of vertices while $E$ is the edge set of the graph. Let the matrix $\mathcal{Y} \in \mathbb{R} ^ {N}$ be the ground-truth class labels corresponding to each vertex of the graph $\mathcal{G}$. 

\begin{figure}[h]
    \begin{center}
    \includegraphics[width=\textwidth]{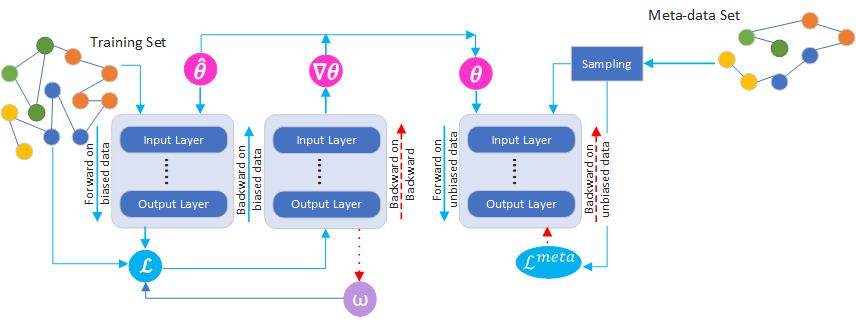}
     \caption{The overall procedure of Meta-GCN's training in a meta-learning manner using a small unbiased meta-data set.}
    \label{framework}
    \end{center}
\end{figure}

Let's assume a graph $\mathcal{G}^{meta} = (V^{meta}, E^{meta}, \mathcal{X}^{meta})$ be meta-graph constructed by an unbiased meta-data set $\mathcal{X}^{meta}$ of size $M$ with the adjacency matrix $A^{meta}$. 
The feature set $\mathcal{X}^{meta}$ is constructed through unbiased sampling from a separate portion of the whole dataset apart from the training and validation set. $V^{meta}$ is a set of vertices where each of its items corresponds to each entry of the set $\mathcal{X}^{meta}$. 
There is an edge $e_{i,j}^{meta}$ in the edge set $E^{meta}$ if and only if ${i, j} \in V^{meta}$.
We represent the ground-truth label set for the meta-data set by $\mathcal{Y}^{meta} \in \mathbb{R} ^ {M}$. 
Given the adjacency matrix $A$ and the feature matrix $\mathcal{X}$, we define a GCN model $f_{\theta}(\mathcal{X}, A)$, where $\theta$ is the set of learnable model parameters. 
Let $\hat{y}=f_{\theta}(\mathcal{X}, A)$ and $\hat{y}^{meta}=f_{\theta}(\mathcal{X}^{meta}, A^{meta})$ be the predicted labels for the training set and meta-data set, respectively. 
Assume that $\mathcal{L}(\hat{\mathcal{Y}}, \mathcal{Y})$ is the loss function for training data, and $\mathcal{L}^{meta}(\hat{\mathcal{Y}}^{meta}, \mathcal{Y}^{meta})$ be that for the meta-data set. 
For the evaluation of this semi-supervised node classification problem, we use the cross-entropy loss function. 
We define an individual losses $l_i(\hat{y}_i, y_i)$ for each training example $i$ and $l^{meta}_j(\hat{y}^{meta}_j, y^{meta}_j)$ for each meta-data example $j$, where $y_i \in \mathcal{Y}$, $y_i^{meta} \in \mathcal{Y}^{meta}$, $\hat{y}_i \in \mathcal{\hat{Y}}$, and $\hat{y}_i^{meta} \in \mathcal{\hat{Y}}^{meta}$. 
We define our problem as minimizing a weighted loss parameterized with $w = \{w_i|1 \leq i \leq N \}$, as 
\begin{equation}
\theta^*(w)=\arg \min_\theta \sum_{i=1}^N w_i l_i(\theta).
\end{equation}
At each step, $\theta$ is updated as estimated weights are updated through:
\begin{equation}
\theta_{t+1}(w)=\theta_{t}(w)-\alpha\nabla\sum_{i=1}^N{w_{i,t} l_i(f_{\theta}(\mathcal{X}, A), y_i)},
\end{equation}

\begin{equation}
\hat{\theta}_{t+1}(\gamma)=\theta_{t}(\gamma)-\alpha\nabla\sum_{i=1}^N{\gamma_{i,t} l_i(f_{\theta}(\mathcal{X}, A), y_i)},
\end{equation}

\begin{equation}
\Tilde{w}_{i,t}=\max(0, -\eta  \displaystyle \frac{\partial} {\partial {\gamma_{i,t}}} \frac{1} {m} \sum_{j=1}^m{l^{meta}_j(f^{meta}_{\hat{\theta}}(\mathcal{X}^{meta}, A^{meta}), y^{meta}_i)}),
\end{equation}

\begin{equation}
\displaystyle w_{i,t} =  \frac{\Tilde{w}_{i,t}}{(\sum_{j=1}^M{\Tilde{w}_{j,t}})+\delta(\sum_{j=1}^M{\Tilde{w}_{j,t}})},
\end{equation}
where $\delta(z)$ is 1 if $z = 0$ and 0 otherwise. 
$\eta$ is the meta-learning rate. $\gamma$ is the weight perturbing parameter. For the GCN model, we define the normalized adjacency matrix as $\Tilde{A}=A+I_N$. 
A degree matrix $\Tilde{D}_{i,i} =  \sum_{j=1}^{N} \Tilde{A}_{i,j}$ can be defined for each node $i$. 
We also calculate the renormalized graph Laplacian $\hat{A}=\Tilde{D}^{-\frac{1}{2}}\Tilde{A}\Tilde{D}^{-\frac{1}{2}}$. 
Considering an n-layer graph-based neural network GCN model for node classification, we compute the output for the layer $l$ as
\begin{equation}
Z^{l}=\sigma(\hat{A}Z^{l-1}{\theta}^{l}),
\end{equation}
where $\sigma(.)$ and ${\theta}^{l}$ are respectively corresponding to the layer's activation function and weights. Finally, we compute the neural networks classifier's results through $f_{\theta}(\mathcal{X}, A)=Simoid(Z^{n})$. Figure \ref{framework} depicts the overall training process of Meta-GCN.

\section{Experiments}
\label{experiments}

To compare the node classification performance of our method with that of the baselines, we ran experiments on two medical datasets, namely Haberman and Diabetes using Meta-GCN and the five other baselines. 
For the SMOTE method, we used the scikit-learn library in Python \cite{pedregosa2011scikit}. For training, validation, testing, and meta-set generation we respectively used 60\%, 10\%, 20\%, and 10\% of the data. For the GraphSMOTE and GCN, we used the same hyperparameters in the publications and the evaluations are done by the source codes shared by the authors. For both oversampling methods, SMOTE and GraphSMOTE, we used an over-sampling scale of 0.8, as it is the best value. 
We adopted two layers GCN model with 32 hidden units. For the activation function of the layers, we used Rectified Linear Unit (ReLU). We summarized the resources for the three metrics, accuracy, macro F1, and AUC-ROC, in Table \ref{tbl:hdcomp}. It is perceivable that our method, Meta-GCN, outperformed all five baselines on all three metrics. Meta-GCN's higher macro F1-score on both datasets suggests that our approach has better performance for minorities as well as the majorities. MLP performed worse than other graph-based baselines which are methods on both datasets for all three metrics. Since SMOTE oversampling method simply performs the interpolation on neighboring nodes without considering the graph structure, its performance is worse than the GCN without oversampling.

For the Diabetes dataset, it is evident that utilizing the graph structure led to substantial improvement. Since GraphSMOTE constructs an embedding space for interpolation and considers the graph structure, its performance is significantly better than SMOTE on this dataset, where much information can be gained from the graph information. However, in terms of accuracy, GraphSMOTE does not perform significantly better than GCN-Weighted. It performs even worse compared to the standard GCN in terms of macro F1 and AUC-ROC.
For Haberman, nodes have less number of features compared to Diabetes. By comparing the results between GCN and GCN-Weighted on macro F1, it can be concluded that the weighting effect is more considerable in this dataset. Judging by insignificant improvement on three metrics it can be inferred that gaining from the graph information in this dataset is limited. Limited information gain from the graph structure can also explain why the improvement of using GraphSMOTE compared to SMOTE is insignificant. However, in this case, Meta-GCN through the use of the meta-set was able to significantly improve the performance on all three different metrics. Considerable improvement of macro F1 by Meta-GCN demonstrates its superior performance in classifying both minority classes and majority classes. Judging by the improvement of macro F1-score on  Diabetes compared to Haberman, which has a higher imbalance ratio, the gain from weighting is more significant.

Based on the results of this experiment, it can be concluded that: 1) For a better comparison and understanding of models none of the metrics are sufficient by themselves. 2) Depending on the dataset the performance can be increased by gaining information from graph structure. 3) Higher performance by the weighting method can be gained with datasets with a higher imbalance ratio. 4) Using the Meta-GCN method significantly improved the classification performance in all three metrics, suggesting that not only improves the overall accuracy of the model in the classification of the examples but also improves the discrimination power of the model for classifying both majorities and minorities.

\begin{table*}[t]
\begin{adjustbox}{width=\textwidth}
        \begin{tabular}{c | c c c | c c c}
         \hline
           &  & Diabetes &  &  & Haberman & \\ [0.25ex] 
         \hline 
         Methods   & Accuracy  & Macro F1 & AUC-ROC & Accuracy & Macro F1 & AUC-ROC \\ [0.25ex] 
        \hline
         MLP & 0.58 ± 0.07 & 0.43 ± 0.07 & 0.55 ± 0.07 & 0.71 ± 0.03 & 0.46 ± 0.04 & 0.45 ± 0.17\\ [0.25ex] 
         GCN & 0.70 ± 0.10 & 0.65 ± 0.09 & 0.72 ± 0.07 & 0.74 ± 0.04 & 0.47 ± 0.05 & 0.55 ± 0.06 \\ [0.25ex] 
         GCN-Weighted & 0.71 ± 0.05 & 0.64 ± 0.24 & 0.66 ± 0.19 & 0.71 ± 0.05 & 0.57 ± 0.05 & 0.46 ± 0.09 \\ [0.25ex] 
         SMOTE & 0.65 & 0.57 & 0.58 & 0.71 & 0.50 & 0.52 \\ [0.25ex] 
         GraphSMOTE & 0.72 ± 0.14 & 0.65 ± 0.09 & 0.70 ± 0.27 & 0.74 ± 0.24 & 0.47 ± 0.05 & 0.54 ± 0.12 \\ [0.25ex] 
         \hline
         Meta-GCN (ours) & \textbf{0.74 ± 0.17} & \textbf{0.70 ± 0.12} & \textbf{0.75 ± 0.15} & \textbf{0.76 ± 0.17} & \textbf{0.65 ± 0.07} & \textbf{0.62 ± 0.09}\\ [0.25ex] 
         \hline
        \end{tabular}
\end{adjustbox}
\caption{Comparison of different methods for imbalanced node classification.}
\label{tbl:hdcomp}
\end{table*}

\section{Conclusion and Future Works}
In this paper, we proposed a meta-learning-based method, name Meta-GCN for dealing with label class imbalance for semi-supervised node classification. The proposed method leverages a small unbiased meta-data set to adaptively learns the example weights through minimization of the meta-data set loss simultaneous to optimizing the model weights. This approach is general purpose and applicable to any graph-structured dataset that suffers from class imbalance. Compared to the other traditional re-weighting methods, Meta-GCN is an end-to-end method and does not require any manual weight setting and extra hyperparameter searching. Our empirical results demonstrate the superiority of Meta-GCN compared to the representative and state-of-the-art approaches in terms of accuracy, macro F1, and AUC-ROC. The comparison of experimental results shows that our model is effective in semi-supervised node classification for graph-structured datasets with the class imbalanced distribution. 
There are several avenues for further investigation. Firstly, in this paper, we proposed a graph-based sampling method that chooses examples for each class with equal probability without considering the graph structure. Therefore, we intend to improve the sampling method for gaining better performance. Second, the proposed method is for node classification and we intend to extend our approach to other applications such as edge prediction or regression tasks.

\printbibliography[heading=subbibintoc]

\end{document}